\documentclass{article}

\usepackage[preprint]{neurips_2024}

\usepackage[utf8]{inputenc}
\usepackage[T1]{fontenc}
\usepackage{microtype}
\usepackage{amsmath,amsfonts}
\usepackage{booktabs}
\usepackage{graphicx}
\usepackage{wrapfig}
\usepackage{multirow,array}
\usepackage{placeins}
\usepackage[dvipsnames]{xcolor}
\usepackage[most,listings]{tcolorbox}
\usepackage{listings}
\newtcolorbox[list inside=prompt,auto counter,number within=section]{prompt}[1][]{
    colbacktitle=black!60,
    coltitle=white,
    fontupper=\footnotesize,
    boxsep=5pt,
    breakable,
    enhanced,
    left=0pt,
    right=0pt,
    top=0pt,
    bottom=0pt,
    boxrule=1pt,
    verbatim,
    #1   
}
\usepackage{hyperref}

\newcommand{\deemph}[1]{\textcolor{gray}{#1}}

\title{Analyzing Information Sharing and \\ Coordination in Multi-Agent Planning}
\author{Tianyue Ou \qquad Saujas Vaduguru \qquad Daniel Fried \\ Language Technologies Institute \\ Carnegie Mellon University \\ \texttt{\{tianyueo,svadugur,dfried\}@andrew.cmu.edu}}

\begin{document}
\maketitle

\begin{abstract}
% (abstract goes here)
% 1, paragraph 1, 2, some of the problems  + contributions: MAS with orchetrator & notebook. 6 sentences. starting 2,3 start with what we are doing.
% last sentence talk about improvements. 
Multi-agent systems (MASs) have pushed the boundaries of large language model (LLM) agents in domains such as web research and software engineering. However, long-horizon, multi-constraint planning tasks involve conditioning on detailed information and satisfying complex interdependent constraints, which can pose a challenge for these systems. In this study, we construct an LLM-based MAS for a travel planning task which is representative of these challenges. We evaluate the impact of a notebook to facilitate information sharing, and evaluate an orchestrator agent to improve coordination in free form conversation between agents. We find that the notebook reduces errors due to hallucinated details by 18\%, while an orchestrator directs the MAS to focus on and further reduce errors by up to 13.5\% within focused sub-areas. Combining both mechanisms achieves a 25\% final pass rate on the TravelPlanner benchmark, a 17.5\% absolute improvement over the single-agent baseline's 7.5\% pass rate. These results highlight the potential of structured information sharing and reflective orchestration as key components in MASs for long horizon planning with LLMs.

\end{abstract}

\section{Introduction}
Large language models have enabled the design of systems that undertake complex tasks in rich, realistic environments like web browsing \citep{Zhou2023WebArenaAR}, software engineering \citep{Yang2024SWEagentAI}, flight booking \citep{Yao2024benchAB}, and general office computer tasks \citep{Xu2024TheAgentCompanyBL}. These systems model a single agent reasoning and taking actions in an environment using tools to solve tasks specified in natural language. While these advances have expanded the abilities of AI systems greatly, solving long-horizon planning tasks, with many interdependent actions, requires reasoning about a number of interdependent constraints, drawing on information from heterogeneous sources that are accessed with different tools, and synthesizing information with accuracy and consistency. Since each action adds information to the agent's context window, and each source of information requires different tools to access, a single agent solving this task needs to reason over increasingly long contexts and choose correctly from a large number of tools. This risks errors and hallucinations, which cascade in long-horizon tasks to result in a failure to fully solve the task. % can we cite some papers for these issues? maybe the TP paper has some analysis

Multi-agent systems (MAS) offer an avenue to tackle these issues. When different agents work together, but each focuses on a part of the task, each agent can reason about less information \citep{hong2024metagptmetaprogrammingmultiagent,dong2024selfcollaborationcodegenerationchatgpt,xia2024agentlessdemystifyingllmbasedsoftware}, critique more effectively  \citep{du2023improvingfactualityreasoninglanguage,tang2024medagentslargelanguagemodels}, explore more solutions \citep{pan2025learningadaptiveparallelreasoning}, and use tools more proficiently. Multi-agent systems have been applied to complex tasks such as web research \citep{anthropic2025multiagent}, stock trading \citep{xiao2025tradingagentsmultiagentsllmfinancial}, software engineering \citep{qian2024chatdevcommunicativeagentssoftware,li2023camel,xia2024agentlessdemystifyingllmbasedsoftware}, and medical decision support \citep{ke2024enhancingdiagnosticaccuracymultiagent,tang2024medagentslargelanguagemodels,kim2024mdagentsadaptivecollaborationllms} to overcome the limitations of single agent systems.

Multiple agents working together require a way to share information between agents, and a way to coordinate how agents act to navigate complex dependencies between different parts of the task. In this paper, we propose approaches to these problems, and systematically evaluate the impact of our solutions in the TravelPlanner benchmark \citep{xie2024travelplannerbenchmarkrealworldplanning}. We use a \emph{notebook} which multiple agents write to and read from to serve as an information sharing mechanism. We also study the benefits of an \emph{orchestrator agent} that reasons about the current state of task completion and chooses how to sequence actions by different agents. We show that this dynamic form of coordination is more effective than a fixed workflow \citep{xiao2025tradingagentsmultiagentsllmfinancial}.

We find that adding a notebook to share information improves performance by 3.75\% in final pass rate over using the dialog alone, with GPT-4o. In a system equipped with a notebook, we find that an orchestrator that reasons about the choice of the next agent improves performance by 1.25\% over a fixed workflow to sequence agents. Overall, these proposed approaches combine to improve performance of a GPT-4o-based system by 5\% over the single-agent baseline, and the same improvement over a multi-agent baseline that does not incorporate these approaches. The final system (that includes both the notebook and an orchestrator agent) based on Claude 4 Sonnet 
% further boosts the improvement with orchestrator to 10\%, and overall 17.5\% 
improves by 17.5\%
over single-agent baseline, further boosting performance.

\section{Proposed Multi-Agent System}
\label{sec:method}

\begin{figure}[t]
  \centering
  \includegraphics[page=1,width=\linewidth]{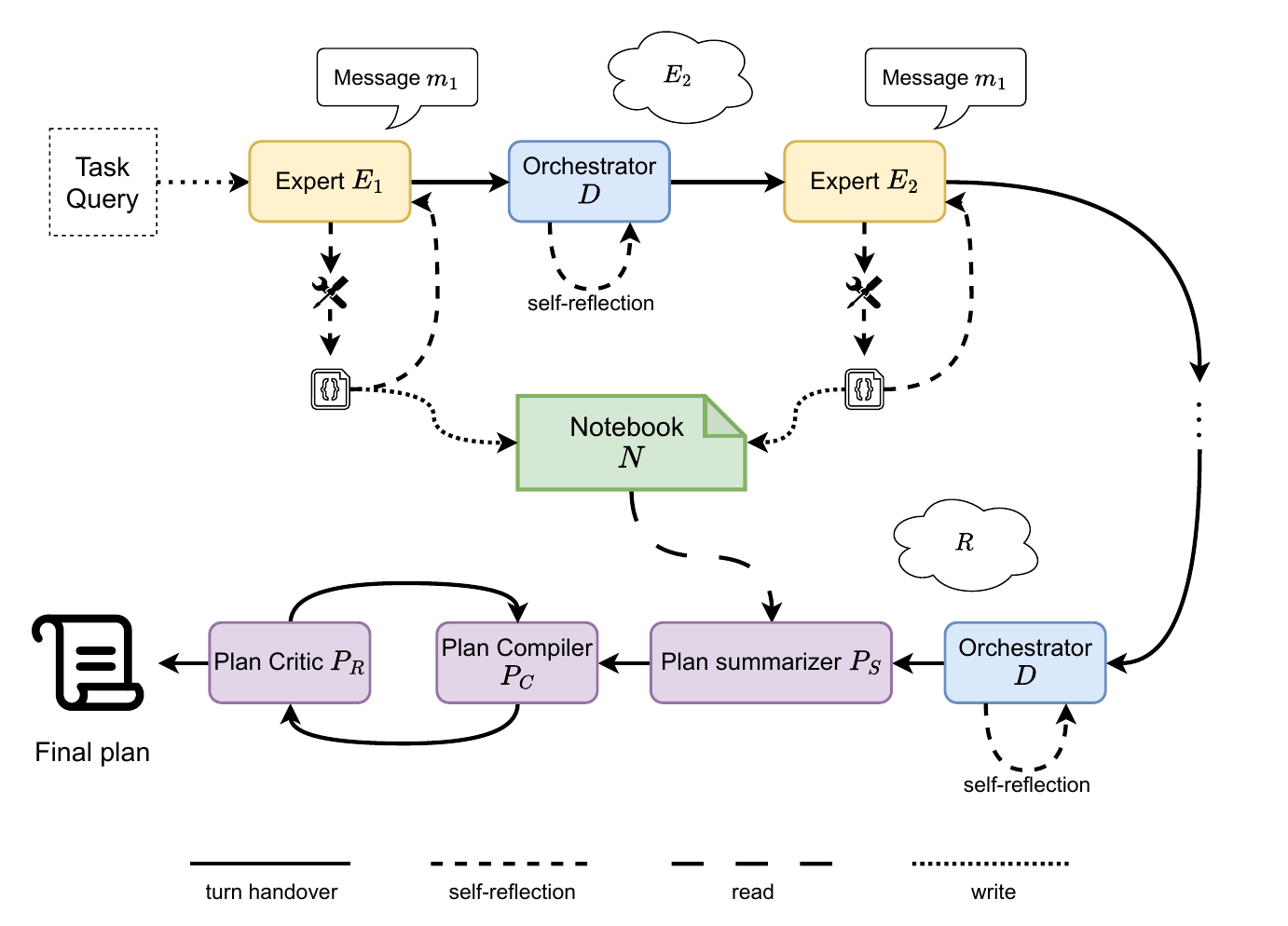}
  \caption{Schematic diagram of our multi-agent system. We implement our multi-agent system using the AutoGen framework \citep{wu2023autogenenablingnextgenllm}.}
  \label{fig:main_figure}
\end{figure}

We propose a multi-agent, LLM-based system for long-horizon planning in tasks that require resolving constraints between specialized sources of information (e.g., travel planning requires finding available hotels, coordinating transportation between them, and choosing restaurants that meet a user's preferences). An overview of our system is shown in \autoref{fig:main_figure}. 

The task is given by a natural language goal $G$. 
% The system consists of \todo{x} main components. We define them as follows:
The system uses a set of LLM-based agents, $\mathcal{A}=\{D, R, P\}\cup\{E_1,\ldots,E_K\}$ where:
\begin{itemize}\setlength{\itemsep}{2pt}
    \item Experts $E_k$: Specialist agents which use tools to retrieve information, such as transportation experts and restaurant experts.
    \item Orchestrator $D$: Decides which agent acts next.
    \item Plan summarizer $P_S$: Surfaces the notebook and task query for plan compilation (not backed by an LLM).
    \item Plan compiler $P_C$: Synthesizes a plan based on the task query and the facts from the notebook.
    \item Plan critic $P_R$: Iteratively refines the plan in conjunction with $P_C$.
\end{itemize}

% We define the conversation space to be a sequence of messages between the agents
% \[
% W \;=\; \{\, [m_1,\ldots,m_\ell] \;:\; \ell \ge 0 \,\}.
% \]
% The state is $w=c\in W$ with public conversational history $c=[m_1,\ldots,m_\ell]$. Each message $m=(u,v)$ has text $u$, and a visibility set $v\subseteq\mathcal{A}$. Each agent $A_i$ has an observation function $o_i:W\to c_i$, the part of conversation visible to agent $A_i$, an action space $Y_i$, and a policy $\pi_i(y\mid o_i(w),G)$. We additionally maintain a Notebook $N$ for structured evidence. The notebook is only visible to a subset of agents. 
% In our implementation, we use the same underlying LLM (e.g. GPT-4o) for all agents, but each agent has a unique prompt for its role.

We define the public conversational history to be the sequence of $\ell$ messages produced by the agents, $c=[m_1,\ldots,m_\ell]$.
We additionally maintain a Notebook $N$ for structured evidence (see \autoref{sec:notebook}), so that the full world state is $w = (c, N, G)$. 
Each agent $A_i$ has an action space $Y_i$, a policy $\pi_i(y\mid o_i(w))$ that uses an observation function $o_i$ to select parts of the state visible to each agent.
The notebook is only visible to a subset of agents. 
In our implementation, we use the same underlying LLM (e.g. GPT-4o) for all expert agents, but each agent has a unique prompt for its role.

% \paragraph{The visibility rules are as follows:}
% \begin{enumerate}\setlength{\itemsep}{2pt}
%     \item \(\forall A_i\in\mathcal{A}:\; o_i(w)\supseteq\{G,\,c\}\).
%     \item For each \(E_k\): \(\tau_k \subseteq o_{E_k}(w)\), where \(\tau_k\) is the tool-call returns. \(\tau_k\) is not appended to \(c\) and is not visible to \(A_j\neq E_k\).
%     \item \(\forall E_k:\; \text{after each tool call, }E_k\text{ writes its returns to }N.\)
%     \item \(o_R(w)\supseteq N\quad\text{and}\quad o_P(w)\supseteq N.\)
%     \item \(o_D(w)=\{G,\,c\};\ \text{$D$ does not read }N\text{ and contributes no public text.}\)
% \end{enumerate}

\paragraph{The action spaces of each agent are:}
\begin{itemize}\setlength{\itemsep}{2pt}
    \item The orchestrator $D$ selects the next agent $A_j$ using only the goal $G$ and the public conversation $c$.
    \item The experts $E_k$ call tools and then write the results to the notebook $N$. The experts also add to the conversation $c$. Tool results are visible only to $E_k$; structured facts must be recorded in $N$.
    \item The plan summarizer $P_S$ prepares a planning brief from $G$ and $N$ and then prompts the plan compiler $P_C$.
    \item The plan compiler $P_C$ produces the final plan as a public message in $c$, which is iteratively refined by continuing the conversation $c$ with the plan critic $P_R$.
\end{itemize}

% \ty{which visibilty section is better} % daniel: I like the natural language one with less notation
\paragraph{The visibility rules are:}
\begin{enumerate}\setlength{\itemsep}{2pt}
    \item All agents observe $G$ and the public conversation $c$.
    \item Experts $E_k$ see the returns of their own tool calls. These returns are not appended to $c$ and not visible to other agents.
    \item Experts automatically write their tool call returns to the notebook $N$. 
    \item The plan summarizer $R$ and planner compiler $P$ read $N$.
    \item The orchestrator $D$ bases decisions only on $(G,c)$; it does not read $N$ and does not contribute to $c$.
\end{enumerate}

\paragraph{Example flow.}
% The orchestrator $D$ chooses the next agent to act. Each expert policy $\pi_{E_k}$ conditions on $(G,c)$ plus its own tool returns to decide when to call tools, summarize the returned values, and persist evidence via the notebook $N$. The plan summarizer $\pi_{P_S}$ consolidates $G$ with $N$ into a planning brief and then hands off to the next agent. The plan compiler $\pi_{P_C}$ conditions on $(c, N, G)$ to produce the final itinerary.

A typical turn of interaction begins with the orchestrator $D$ scanning the conversation history $c$ and the stated goal $G$ to identify the current issue to solve. $D$ grants control to an expert $E_i$. $E_i$ issues one or more tool calls to query available options; the raw tool call results are visible only to $E_1$. From these, $E_i$ responds with its part of the solution to $c$, and at the same time writes the results of its tool calls to $N$. Observing the updated conversation, $D$ selects another expert $E_j$, which repeats the pattern: private tool calls, then adding to the conversation $c$ and notebook $N$. No expert ever sees another expert’s raw tool call results. When $D$ judges that sufficient evidence has been accumulated across domains, it selects the plan summarizer $P_S$ to act. $P_S$ uses the original goal $G$ and the contents of the notebook $N$ to assemble a coherent planning brief and passes this brief to $P_C$. Finally, $P_C$ reads $(G,c,N)$ and produces the final itinerary. Throughout, $D$ itself does not contribute public text and never reads $N$.

\section{TravelPlanner}
We use TravelPlanner \citep{xie2024travelplannerbenchmarkrealworldplanning} as the environment to evaluate our multi-agent systems. TravelPlanner features long-horizon travel plan generation with agents. 
% As a closed, tool-based sandbox for long horizon travel itinerary planning, it emphasizes multi-constraint long horizon planning. 
Agents need to retrieve flight, hotel, resturants information by calling tools on a database of around four million entries. Queries in the TravelPlanner benchmark ask the agent to make three, five, and seven day plans. Each query is accompanied by two types of constraints -- \emph{hard} user-specified constraints explicitly provided in the query, and \emph{commonsense} constraints imposed by general real-world factors.

\subsection{Metrics}
The TravelPlanner benchmark measures model performance using  three types of metrics. The primary metrics are macro pass rates: \textbf{Commonsense Macro Pass Rate} and \textbf{Hard Macro Pass Rate}, which measure the fraction of tasks where the system was able to produce a plan that satisfied \emph{all} commonsense constraints and \emph{all} user-specified (hard) constraints, respectively. \textbf{Final Pass Rate} combines these, measuring the fraction of tasks where the system produced a plan that satisfies all constraints. To give a finer-grained evaluation, \textbf{Commonsense Micro Pass Rate} and \textbf{Hard Micro Pass Rate} measure the fraction of constraints of that type which were passed, and \textbf{Delivery Rate} measures the fraction of tasks where the agent delivered a final plan within the step limit. 
% \ty{too many metrics -> introduce the main metrics.  divide the table into micro rate, macro rate, final pass rate.}

\section{Role of the Notebook in Mitigating Hallucinations}
\label{sec:notebook}

\begin{table}[h]
\centering

% \begin{tabular}{lcc}
% \toprule
% Metric & With Notebook & Without Notebook \\
% \midrule
% \textbf{Final Pass Rate (\%)}        & \textbf{5.00}  & 1.25  \\
% % \midrule
% \textbf{Commonsense Macro Pass (\%)} & \textbf{12.50} & 6.25  \\
% \textbf{Hard Macro Pass Rate (\%)}   & \textbf{13.75} & 5.00  \\
% \midrule
% Delivery Rate (\%)                   & 96.25 & 97.50 \\
% Commonsense Micro Pass (\%)          & 71.41 & 67.81 \\
% Hard Micro Pass Rate (\%)            & 39.78 & 12.90 \\
% \bottomrule
% \end{tabular}
\begin{tabular}{lcc}
\toprule
Metric & Without Notebook & With Notebook \\
\midrule
% \midrule
Commonsense Macro Pass (\%) & 6.25 & \textbf{12.50} \\
Hard Macro Pass Rate (\%) & 5.00 & \textbf{13.75} \\
Final Pass Rate (\%) & 1.25 & \textbf{5.00} \\
\midrule
\deemph{Delivery Rate (\%)} & \deemph{97.50} & \deemph{96.25} \\
\deemph{Commonsense Micro Pass (\%)} & \deemph{67.81} & \deemph{71.41} \\
\deemph{Hard Micro Pass Rate (\%)} & \deemph{12.90} & \deemph{39.78} \\
\bottomrule
\end{tabular}
\vspace{2mm}
\caption{Comparison of our multi-agent system with and without the notebook, using GPT-4o as the base LLM.}
\label{tab:rq1_table_reorganized}
\end{table}

Planning for complex long-horizon tasks, such as multi-day travels, involves managing a large number of details. It is crucial for multi-agent planning systems to preserve all the information gathered by different agents along the way and present it accurately without hallucination in the end. Yet, it remains a challenge for current agents to maintain precise details, such as the names of places, restaurants, and hotels after long multi-turn conversations. We observed that 17.5\% of generated plans have at least one entity that was hallucinated (i.e., is not a part of the sandbox environment) before the addition of the notebook.

\begin{wrapfigure}{r}{0.5\linewidth} 
    \vspace{-10mm}
    \centering
    \vspace{5pt} % adjust vertical space if needed
    \includegraphics[width=\linewidth]{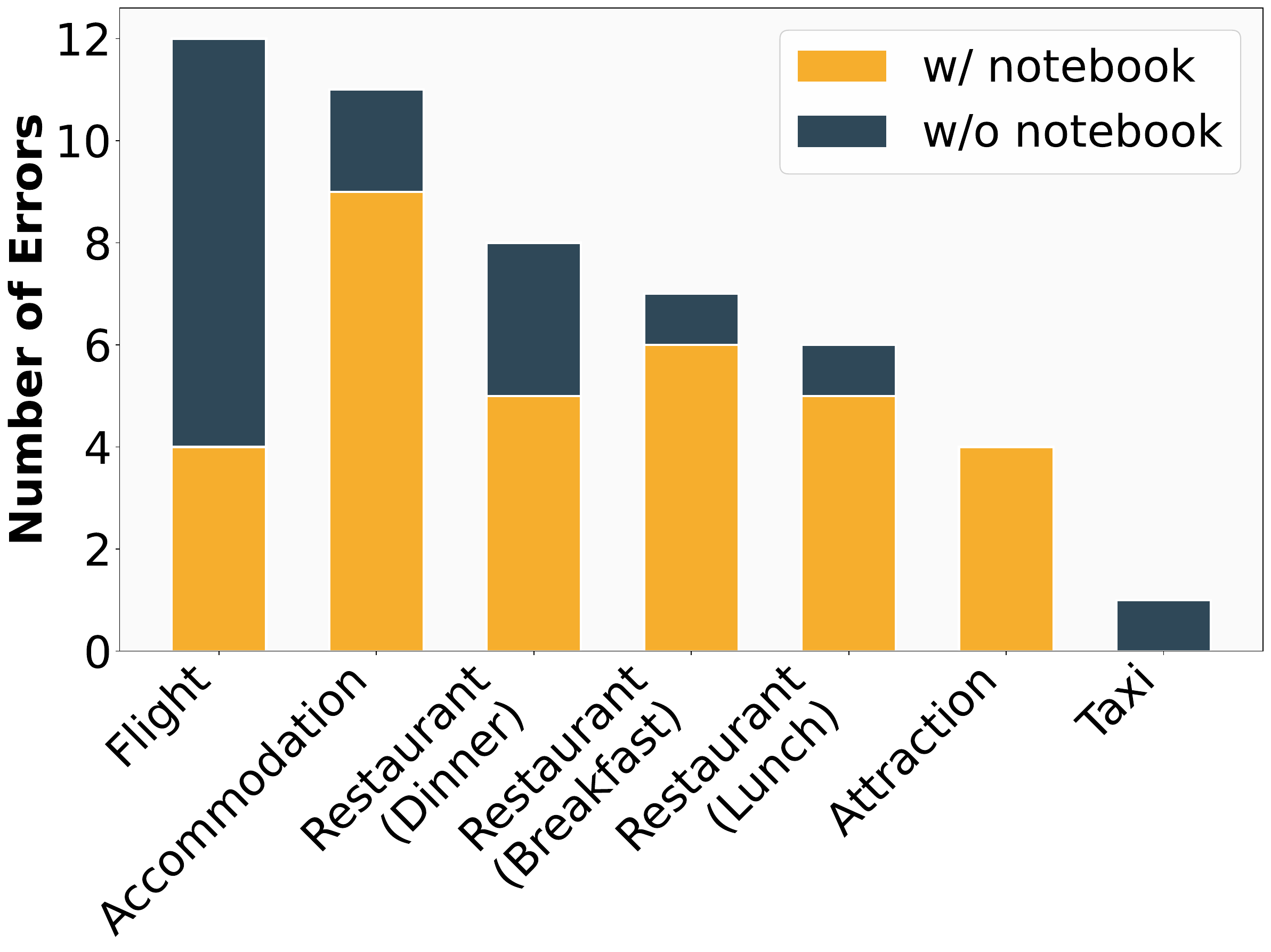}
    \vspace{-6mm}
    \caption{Number of 
    % \texttt{is\_valid\_information\_in\_the\_sandbox}
    errors where a model references information not in the sandbox before and after the introduction of a notebook. Results are for a GPT-4o-based system. 
    % Access to a notebook gives agents a structured means to communicate with other agents and reduces hallucinations of information not in the sandbox based only on the messages exchanged.
    }
    \label{fig:hallucination}
    % \vspace{-25mm}
    \vspace{-6mm}
\end{wrapfigure}

In our first research question, we examine the effectiveness of using a notebook as a mitigation strategy against hallucinations in multi-agent planning. We instantiate the notebook mechanism defined in \autoref{sec:method} as a grounding constraint on the final plan. Each expert $E_k$ writes the returns of its private tool calls to $N$, which is visible to only the planner agents $P_S$ and $P_C$. For planning, $P_S$ prompts $P_C$ with all the previous conversation and the notebook, and the planner $P_C$ conditions on $(c, N, G)$ to produce the final itinerary.

% \subsection{Experiment}
% In building up our MAS, we start with a simple workflow schema where agents speak in fixed order and does not go back once it ends its turn. 
We evaluate the effect of the notebook by comparing it to a system that uses only the public conversational history $c$ in generating the plan. In both these systems, a simple workflow determines the fixed order in which agents speak.
The order of speaking is transportation agent, hotel agent, restaurant agent, attraction agent, plan summarizer, and plan compiler. 
In the notebook setting, the notebook captures all results from retrieval tool calls by the expert agents. This notebook is not part of the conversation but is made available to the plan compiler through the plan summarizer. 
% We evaluated both settings in our TravelPlanner evaluation set.

% \subsection{Notebook helps retain the right details}

We find that the notebook helps improve MAS performance substantially, with an overall increase of 3.75\% in final pass rate with GPT-4o, as shown in \autoref{tab:rq1_table_reorganized}. We observe improvements in both macro and micro constraints pass rate, showing the notebook's effectiveness in helping our multi-agent system to deliver travel plans that meet all constraints. %both make better commonsense and more costumed to specific user requirements.

To assess how much the notebook helps reduce hallucination, we evaluated the number of times the MAS returned the name of a place that is not in the sandbox environment. This is a specific category of errors captured by the metric of \texttt{is\_valid\_information\_in\_the\_sandbox}. \autoref{fig:hallucination} shows the number of errors of this type. We can see the number of errors reduces substantially when the notebook is introduced. In particular, for flight numbers, our multi-agent system eliminates more than half the errors. With the help of notebooks, agents can verify the exact name or numbers of places and tickets and avoid hallucinating them in long conversations. %We show some examples in \autoref{app:}

\section{Orchestrator-led Conversation as an Alternative to Workflows}
\begin{figure}[t]
    \centering
    \includegraphics[width=0.8\linewidth]{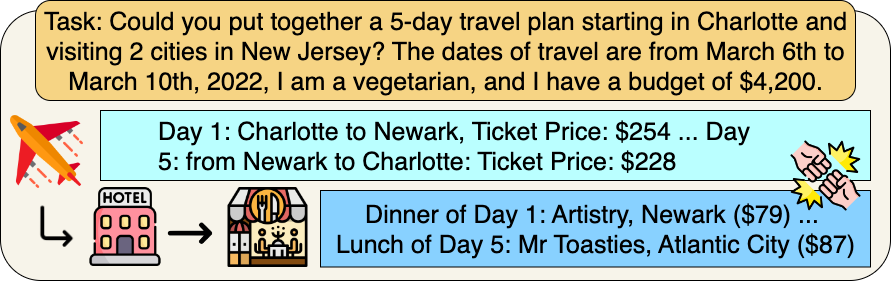}
    \caption{Example of a conflict involving interdependent constraints.}
    \label{fig:free_chat_motivation}
    \vspace{-0.5em}
    \vspace{-5mm}

\end{figure}

Complex long horizon planning tasks are challenging with their interdependent constraints: in order to satisfy later constraints, one may need to go back and revise previous plans. An example is illustrated in \autoref{fig:free_chat_motivation}, where a later decision to have dinner at Mr Toasties pushes the total budget over the limit. However, Mr Toasties is the only restaurant that offers vegetarian options. Resolving the budget problem requires re-visiting previous options, which may involve selecting cheaper flight options. A fixed workflow MAS, where agents act in a pre-determined order, may not have the flexibility to revisit previous steps. On the other hand, an effective MAS for long-horizon tasks should have the freedom to choose where to spend the most effort, and be able to jump around to different parts of the plan to resolve interdependent constraints. We investigate \emph{orchestrator-led conversation}, where an LLM agent chooses which expert agent goes next. We enable the orchestrator to engage in self-reflection \citep{shinn2023reflexion} that allows it to reason more effectively about its choice.
% We study a free-conversation format in MAS with a self-reflective chat manager to orchestrate the conversation. 

\begin{table}[h]
\centering

% \begin{tabular}{lcccc}
% \toprule
%  & \multicolumn{2}{c}{Critique \& Free conversation} & \multicolumn{2}{c}{Fixed order workflow} \\
% \cmidrule(lr){2-3} \cmidrule(lr){4-5}
% Metric & GPT-4o & Claude Sonnet 4 & GPT-4o & Claude Sonnet 4 \\
% \midrule
% \textbf{Final Pass Rate (\%)}            & \textbf{6.25}  & \textbf{25.00} & \textbf{5.00}  & \textbf{15.00} \\
% \midrule
% \textbf{Commonsense Macro Pass (\%)}     & \textbf{15.00} & \textbf{32.50} & \textbf{12.50} & \textbf{31.25} \\
% \textbf{Hard Macro Pass Rate (\%)}       & \textbf{13.75} & \textbf{33.75} & \textbf{13.75} & \textbf{25.00} \\
% \midrule
% Delivery Rate (\%)                       & 83.75 & 87.50 & 96.25 & 96.25 \\
% Commonsense Micro Pass (\%)              & 65.00 & 73.75 & 71.41 & 77.03 \\
% Hard Micro Pass Rate (\%)                & 24.73 & 69.89 & 39.78 & 63.44 \\
% \bottomrule
% \end{tabular}
\begin{tabular}{lcccc}
\toprule
& &  & \multicolumn{2}{c}{Self-Reflection \&} \\ 
 & \multicolumn{2}{c}{Fixed Order Workflow} & \multicolumn{2}{c}{Orchestrator-led Conversation} \\
\cmidrule(lr){2-3} \cmidrule(lr){4-5}
Metric & GPT-4o & Claude Sonnet 4 & GPT-4o & Claude Sonnet 4 \\
\midrule
Commonsense Macro Pass (\%) & 12.50 & 31.25 & 15.00 & \textbf{32.50} \\
Hard Macro Pass Rate (\%) & 13.75 & 25.00 & 13.75 & \textbf{33.75} \\
Final Pass Rate (\%) & 5.00 & 15.00 & 6.25 & \textbf{25.00} \\
\midrule
\deemph{Delivery Rate (\%)} & \deemph{96.25} & \deemph{96.25} & \deemph{83.75} & \deemph{87.50} \\
\deemph{Commonsense Micro Pass (\%)} & \deemph{71.41} & \deemph{77.03} & \deemph{65.00} & \deemph{73.75} \\
\deemph{Hard Micro Pass Rate (\%)} & \deemph{39.78} & \deemph{63.44} & \deemph{24.73} & \deemph{69.89} \\
\bottomrule
\end{tabular}
\vspace{2mm}
\caption{Comparison between fixed order workflow MAS and orchestrator-led conversation MAS on both GPT-4o and Claude Sonnet 4.}
\label{tab:rq2-metrics}
\end{table}

% \subsection{Experiment}
% We evaluated how increased in flexibility allows our multi-agent planning system to have freedom in focusing on the part of the plans and come up with overall result improvement. 
% We evaluated the models in the same TravelPlanner setting. 
% In the free conversation system, we designed an orchestrator that organizes the conversation. 
% Our orchestrator $D$ looks at the entire conversation so far and the overall objective $G$, then selects the next agent to speak. It also contains a self-reflective section in its output to help guide its reasoning in orchestrating the conversation. 

% \subsection{Results}

% \begin{wrapfigure}{r}{0.4\linewidth} % 'r' for right, 'l' for left
%     \centering
%     \vspace{-10pt} % adjust vertical space if needed
%     \includegraphics[width=\linewidth]{figures/agent_revisits_v5.pdf}
%     \caption{\todo{Your descriptive caption here.}}
%     \label{fig:focus}
%     \vspace{-5pt} % adjust to control spacing after figure
% \end{wrapfigure}

We compare the performance of the systems that use a fixed workflow (without the orchestrator) and the self-reflection and free conversation (enabled by the orchestrator) in \autoref{tab:rq2-metrics}. Despite a drop in overall delivery rate due to conversation cut-off, the final pass rate of our orchestrator-led conversation system with self-reflection still improves by 1.25\% for GPT-4o and 10\% for Claude Sonnet 4. It shows that our MAS benefits from allowing for free conversations led by an orchestrator. We further see that the macro pass rate of our orchestrator-led conversation system increases despite decreases in micro pass rate. This indicates that a single travel plan made by our system is more likely to have all of the different constraints satisfied for a single task, reflecting a better resolution of the complex interdependence between different constraints. On the other hand, even though the fixed workflow baseline has more constraints satisfied overall, these satisfied constraints are not coordinated well within a single travel plan. Satisfying some of them may result in the violation of others on the same travel plan, without improving the overall macro pass rate.

\begin{table}[h]
\centering
\begin{tabular}{lcccc}
\toprule
 & Transportation & Hotel & Attraction & Restaurant \\
\midrule
\multicolumn{5}{l}{\textbf{Average \# of revisits per-task}} \\
\midrule
Fixed Order Workflow & \multicolumn{4}{c}{\emph{Fixed order workflow has no revisits}} \\
\addlinespace[4pt]
Orchestrator-led Conversation & 0.66 & 0.63 & 0.08 & 0.03 \\
\midrule
\multicolumn{5}{l}{\textbf{Constraints Failed (\%)}} \\
\midrule
Fixed Order Workflow (\%) & 18.18 & 24.53 & 1.32 & 3.92 \\
Orchestrator-led Conversation (\%) & 13.24 & 11.03 & 1.49 & 6.30 \\
$\Delta$ Improvement (\%) & \textcolor{ForestGreen}{\(\downarrow\) 4.94} & \textcolor{ForestGreen}{\(\downarrow\) 13.50} & \textcolor{red}{\(\uparrow\) 0.17} & \textcolor{red}{\(\uparrow\) 2.38} \\
\bottomrule
\end{tabular}
\vspace{2mm}
\caption{The top two rows are number of revisits to each expert agent, averaged over tasks. The bottom two rows show the percentage of failed constraints in each expert agent's respective domain.}
\label{tab:revisits_errors}
\end{table}

\paragraph{Orchestrator-led conversation reduces errors with its flexible focus.}
We conduct an analysis which shows that our orchestrator-led conversation system can identify and select its areas to focus on and reduce errors in these areas. We quantify focus shift by measuring frequency and destination of revisits. A revisit to an expert agent happens when the expert agent has spoken already but the conversation was passed back to it to speak again. We categorize constraints into the domain areas of the experts (e.g., transportation, hotel) and compute the percentage of constraints failed in each category. (See \autoref{app:rq2_categories} for the categorizations.) As shown in \autoref{tab:revisits_errors}, the transportation and hotel agent are the top two destinations of revisits, averaging at 0.66 and 0.63 revisits per task. Correspondingly, the failure rates of constraints in these two areas have dropped substantially when compared to the fixed order workflow system from \autoref{sec:notebook}. %The error categories of each area are detailed in \autoref{app:rq2_categories}.

% The top origins of focus switch are hotel followed by transportation, and the top destinations of focus switch are transportation and hotel respectively. This indicates that our free chat system chooses to spent more flexible effort into coordinating transportation and hotels. However, transportation and hotel are not the only area that requires more focus, satisfying constraints on restaurant and attractions also require frequent re-vising of transportation and hotel plans, making them a significant origin of focus shift too.

% \paragraph{Critique helps orchestrator to lead conversation}

\section{Multi-agent System vs. Single Agent}

\begin{table}[h]
\centering
% \begin{tabular}{lcccc}
% \toprule
% \multirow{2}{*}{Metric} & \multicolumn{2}{c}{Multi-Agent (Critique)} & \multicolumn{2}{c}{Single-Agent Baseline} \\
% \cmidrule(lr){2-3}\cmidrule(lr){4-5}
%  & GPT-4o & Claude 4 Sonnet & GPT-4o & Claude 4 Sonnet \\
% \midrule
% \textbf{Final Pass Rate (\%)}            & \textbf{6.25}  & \textbf{25.00} & \textbf{1.25}  & \textbf{7.50}  \\
% \midrule
% \textbf{Commonsense Macro Pass (\%)}     & \textbf{15.00} & \textbf{32.50} & \textbf{3.75}  & \textbf{17.50} \\
% \textbf{Hard Macro Pass Rate (\%)}       & \textbf{13.75} & \textbf{33.75} & \textbf{5.00}  & \textbf{13.75} \\
% \midrule
% Delivery Rate (\%)                       & 83.75 & 87.50 & 91.25 & 91.25 \\
% Commonsense Micro Pass (\%)              & 65.00 & 73.75 & 68.59 & 67.50 \\
% Hard Micro Pass Rate (\%)                & 24.73 & 69.89 & 18.28 & 34.41 \\
% \bottomrule
% \end{tabular}
\begin{tabular}{lcccc}
\toprule
\multirow{2}{*}{Metric} & \multicolumn{2}{c}{Single-Agent} & \multicolumn{2}{c}{MAS} \\
\cmidrule(lr){2-3}\cmidrule(lr){4-5}
 & GPT-4o & Claude 4 Sonnet & GPT-4o & Claude 4 Sonnet \\
\midrule
Final Pass Rate (\%) & 1.25 & 7.50 & 6.25 & \textbf{25.00} \\
\midrule
Commonsense Macro Pass (\%) & 3.75 & 17.50 & 15.00 & \textbf{32.50} \\
Hard Macro Pass Rate (\%) & 5.00 & 13.75 & 13.75 & \textbf{33.75} \\
\midrule
\deemph{Delivery Rate (\%)} & \deemph{91.25} & \deemph{91.25} & \deemph{83.75} & \deemph{87.50} \\
\deemph{Commonsense Micro Pass (\%)} & \deemph{68.59} & \deemph{67.50} & \deemph{65.00} & \deemph{73.75} \\
\deemph{Hard Micro Pass Rate (\%)} & \deemph{18.28} & \deemph{34.41} & \deemph{24.73} & \deemph{69.89} \\
\bottomrule
\end{tabular}
\vspace{2mm}
\caption{Comparison between the original single-agent implementation in TravelPlanner and our multi-agent system. The multi-agent system improves substantially over the single-agent system.}
\label{tab:critique-vs-single}
\end{table}
Finally, we compare our final MAS system using the notebook and orchestrator-led conversation and self-reflection against the single-agent baseline approach from \citet{xie2024travelplannerbenchmarkrealworldplanning} in \autoref{tab:critique-vs-single}. 
% \daniel{we never defined the single-agent baseline, did we? Is this from the original paper?}\ty{it is from original paper}
The MAS consistently outperforms the single-agent setup in terms of final pass rate, showing an absolute improvement of 5\% with GPT-4o and 17.5\% with Claude Sonnet 4. This demonstrates that, when the system succeeds in delivering a plan, the results are of significantly higher quality, satisfying more of the complex interdependent constraints.

However, the MAS also exhibits a drop in delivery rate compared to the single-agent baseline, from 91.25\% to 83.75\% for GPT-4o and from 91.25\% to 87.5\% for Claude Sonnet 4. This reflects the higher difficulty of coordination in a multi-agent setting: the system is often ``trying harder'' to resolve conflicting constraints and more selective in what it considers an acceptable plan. When it does succeed in delivering, the outputs are substantially more reliable, but the added complexity can increase the chance of non-delivery. Nevertheless, the final pass rates are consistently higher with the multi-agent system in comparison to the single-agent baseline.

\section{Conclusion}
We analyzed how structured information sharing and coordination in multi-agent LLM systems enable improvements in long-horizon, multi-constraint planning. 
Our multi-agent system uses a persistent notebook to allow agents to condition on relevant information and avoid hallucinations, and an orchestrator agent to direct conversation between the agents that enables the system to iteratively resolve constraints.
Controlled experiments show the benefits of both of these components, and allow the system to outperform a single-agent baseline.
% Through controlled experiments, we step by step built a MAS by combining knowledge persistent notebook and orchestrator-led free conversation. 
% We further show how such system outperforms our single-agent baseline and analyze how these mechanisms individually contribute to plan generation. 
By studying long-horizon planning through the lens of information sharing and coordination, our study identifies two practical levers: grounded memory and flexible dialog for building multi-agent systems that more reliably satisfy complex, interdependent constraints while maintaining accuracy over extended reasoning horizons.

These results highlight both the promise and the remaining gap: coordination introduces overhead that can reduce delivery under strict step limits, and resolving coupled constraints still requires better conflict detection. Closing these gaps will further benefit multi-agent systems in planning-related domains.

% system makes more efforts to try to resolve, but do not always deliver. when it dleivers, 

% table that counts numebr of rounds

% fork autogen, have a fork autogen, and then add that in the pip requirements

\section*{Acknowledgments}

This work was supported by a grant from the Defence Science and Technology Agency.
We are grateful to Shearman Chua, Zhiqian Song, Jonathan Tan, Marcus Loke, Shan Jie Yong, Graham Neubig, Zaid Sheikh, and Yueqi Song for helpful feedback on this work.

% ---- Bibliography (place these two lines before \end{document}) ----
\bibliographystyle{plainnat}
\bibliography{reference}

\clearpage
\appendix

\clearpage % start a fresh page so the section + table stay together
\section{Mapping of Error Categories}
\label{app:rq2_categories}

\begin{center}
\footnotesize
\label{tab:category-area-mapping}

\setlength{\tabcolsep}{2.5pt}
\renewcommand{\arraystretch}{1.22}
\begin{tabular}{|>{\raggedright\arraybackslash}p{0.52\linewidth}|>{\raggedright\arraybackslash}p{0.18\linewidth}|}
\hline
\textbf{Validation Category} & \textbf{Area} \\
\hline
Accommodation Rules & \multirow[t]{5}{*}{Hotel} \\
\cline{1-1}
City Valid - Accommodation & \\
\cline{1-1}
Room Rule Compliance & \\
\cline{1-1}
Room Type Preferences & \\
\cline{1-1}
Budget/Cost Compliance & \\
\hline
City Valid - Restaurant (Breakfast/Lunch/Dinner) & \multirow[t]{3}{*}{Restaurant} \\
\cline{1-1}
Diverse Restaurants & \\
\cline{1-1}
Cuisine Preferences & \\
\hline
City Valid - Attraction & \multirow[t]{5}{*}{Attraction} \\
\cline{1-1}
Diverse Attractions & \\
\cline{1-1}
Within Current City & \\
\cline{1-1}
Within Sandbox (No Hallucination) & \\
\cline{1-1}
Complete Information & \\
\hline
City Valid - Transportation & \multirow[t]{3}{*}{Transportation} \\
\cline{1-1}
Reasonable City Route & \\
\cline{1-1}
Transportation Consistency & \\
\hline
\end{tabular}

\end{center}
\clearpage
\section{Prompts}
\begin{prompt}[title={System Prompt For Transportation Expert Agent.}, listing only, listing engine=listings, breakable]
You are a helpful assistant. You are cooperating with others to plan a trip.
You are responsible for planning the transportation only.
Decide on the dates and cities and get both the departure and return flights.
You don't have to decide on the right cities on the first try; adjust them if needed.

Call flight\_search(Departure City, Destination City, Date):
Description: A flight information retrieval tool.
Parameters:
  Departure City: The city you'll be flying out from.
  Destination City: The city you aim to reach.
  Date: The date of your travel in YYYY-MM-DD format.
Example:
  flight\_search(New York, London, 2022-10-01) would fetch flights
  from New York to London on October 1, 2022.
\end{prompt}
\vspace{20mm}

\begin{prompt}[title={System Prompt For Hotel Expert Agent.}, listing only, listing engine=listings, breakable]
Your are a helpful assistant. You are cooperating with others to plan a trip. You are responsible for making a plan for the hotels only. Hotel plan should be in the form of day x, hotel x. You are returning home on the last day, so do not plan hotel for the last day. You must state how many nights of hotel is needed at the beginning. State the names exactly as they appear, do not abbreviate or replace them with other phrases. Call hotel\_search(city):
Description: Discover accommodations in your desired city.
Parameter: City - The name of the city where you're seeking accommodation.
Example: hotel\_search(Rome) would present a list of hotel rooms in Rome
The returned list also include requirements by each hotel. You must explicitly check the plan against each requirements by hotels. Mark your checking process by <checking hotel requirements> ... </checking hotel requirements> 
\end{prompt}
\vspace{20mm}

\begin{prompt}[title={System Prompt For Attraction Expert Agent.}, listing only, listing engine=listings, breakable]
Your are a helpful assistant. You are cooperating with others to plan a trip. You are responsible for making a plan for the attractions only. State the names exactly as they appear, do not abbreviate or replace them with other phrases. Call attraction\_search(City):
Description: Find attractions in a city of your choice.
Parameter: City – The name of the city where you're seeking attractions.
Example: attraction\_search(London) would return attractions in London.
\end{prompt}
\vspace{20mm}

\begin{prompt}[title={System Prompt For Resturant Expert Agent.}, listing only, listing engine=listings, breakable]
Your are a helpful assistant. You are cooperating with others to plan a trip. You are responsible for making a plan on the resturants only. You should pick different resturants for the meals. State the names exactly as they appear, do not abbreviate or replace names with other phrases. Call resturant\_search(City):
Description: Explore dining options in a city of your choice.
Parameter: City – The name of the city where you're seeking restaurants.
Example: resturant\_search(Tokyo) would show a curated list of restaurants in Tokyo.
\end{prompt}

\end{document}